\documentclass[runningheads]{llncs}
\usepackage[T1]{fontenc}
\usepackage{amsmath}
\usepackage{amssymb}
\usepackage{booktabs}
\usepackage{multirow}
\usepackage{graphicx}
\usepackage{array}
\usepackage{caption}
\usepackage{subcaption}
\usepackage{subfig}
\usepackage{rotating}
\usepackage{adjustbox}
\usepackage[dvipsnames]{xcolor}

\begin{document}
\title{LLMs-based Augmentation for Domain Adaptation in Long-tailed Food Datasets}
\titlerunning{LLMs-based Augmentation for Domain Adaptation Long-tailed}
%

\author{Qing Wang,
Chong-Wah Ngo, Ee-Peng Lim, Qianru Sun }

\authorrunning{Q. Wang et al.}
\institute{
School of Computing and Information Systems\\
Singapore Management University \\ 
Singapore \\
\email{qingwang.2020@phdcs.smu.edu.sg} \\
\email{\{cwngo, eplim, qianrusun\}@smu.edu.sg}
}

\maketitle              

\begin{abstract}
Training a model for food recognition is challenging because the training samples, which are typically crawled from the Internet, are visually different from the pictures captured by users in the free-living environment. In addition to this domain-shift problem, the real-world food datasets tend to be long-tailed distributed and some dishes of different categories exhibit subtle variations that are difficult to distinguish visually. In this paper, we present a framework empowered with large language models  (LLMs) to address these challenges in food recognition. We first leverage LLMs to parse food images to generate food titles and ingredients. Then, we project the generated texts and food images from different domains to a shared embedding space to maximize the pair similarities. Finally, we take the aligned features of both modalities for recognition. With this simple framework, we show that our proposed approach can outperform the existing approaches tailored for long-tailed data distribution, domain adaptation, and fine-grained classification, respectively, on two food datasets.


\end{abstract}
\section{Introduction}
\begin{figure}[t!]
\centering
\begin{subfigure}{.28\textwidth}
  \centering
  \includegraphics[width=0.85\linewidth]{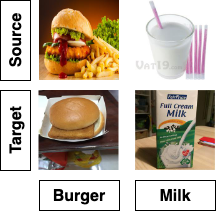}
  \caption{Domain shifts}
  \label{fig_da_samples} 
\end{subfigure}%
\begin{subfigure}{.4\textwidth}
  \centering
  \includegraphics[width=0.85\linewidth]{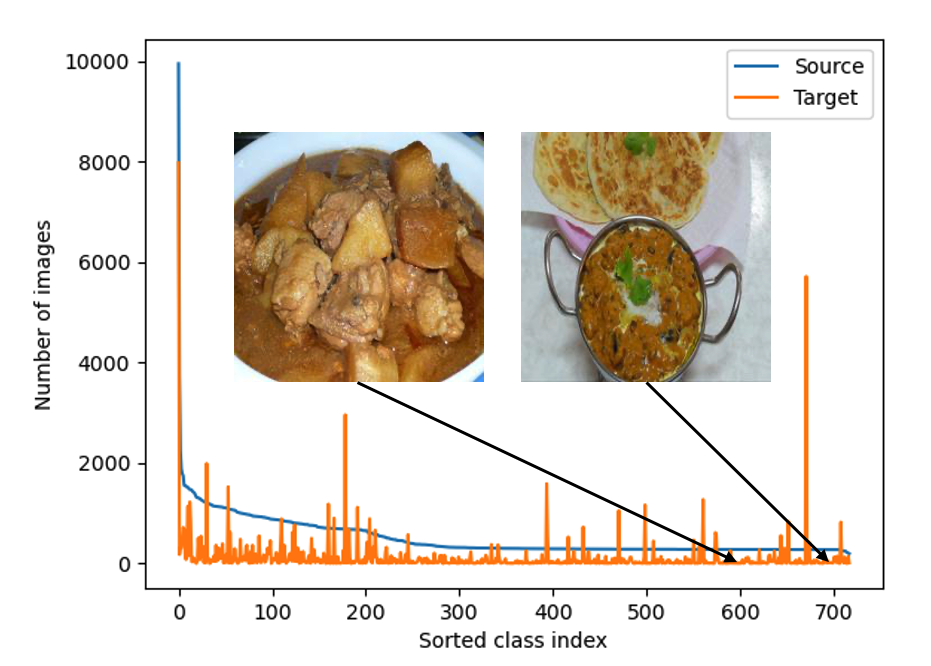}
  \caption{Distribution imbalance}
  \label{fig:dist}
\end{subfigure}%
\begin{subfigure}{.25\textwidth}
  \centering
  \includegraphics[width=0.8\linewidth]{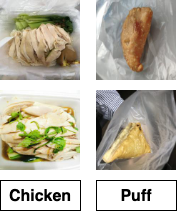}
  \caption{Fine-grained}
  \label{fig_fg_samples} 
\end{subfigure}
\caption{The FoodAI-HPB dataset highlights the challenges in real-world food datasets. (a) Domain shifts: a noticeable disparity exists between the source and target domains concerning factors such as background, lighting, food shapes, and serving styles. (b) Distribution imbalance: both domains exhibit an imbalanced nature with distinct categorical distributions. For example, some classes (e.g., chicken pongteh and makhani dal shown in the plot) suffer from being under-represented in both the source and target domains. (c) Fine-grained categorization: food classes that are highly similar but differ in their ingredient compositions. The chicken rice in the source domain is visually similar to the boiled kampung chicken in the target domain. Similarly, the curry puff and tuna puff resemble each other despite belonging to different categories.} 
\label{fig:foodaihpb-exp-dist}
\end{figure}

Food recognition is a fundamental task to automate dietary assessment. Daily food logging helps to monitor our dietary habits and make informed decisions about nutrition balance. However, food recognition is challenged by the intricate nature of real-world food datasets. Fig.~\ref{fig:foodaihpb-exp-dist} shows a real-world dataset (FoodAI-HPB), where the images in the source domain are crawled from the Internet while the images in the target domain are captured in the free-living environment and uploaded by residents through a mobile app. Fig.~\ref{fig_da_samples} highlights the discrepancy between the source and target images (domain shift) while the uneven distributions of images across food categories can be observed from  Fig.~\ref{fig:dist} (long-tailed distribution). Furthermore, there exist visually and semantically similar classes in the dataset as shown in Fig.~\ref{fig_fg_samples}, challenging fine-grained food categorization.

To address these open issues in the real-world datasets, an increasing number of works have been proposed for domain gaps~\cite{ganin2016domain,zhang2023free,zhou2024unsupervised}, data imbalance~\cite{kang2019decoupling,wang2024unified}, and fine-grained recognition~\cite{liang2022penalizing,min2023large,rodenas2022learning}. Recently, there are also research attempts~\cite{gu2022tackling,yang2022multi} to tackle long-tailed recognition under domain shifts. In~\cite{yang2022multi}, samples from the same category but different domains are treated as positive, while cross-class samples are considered negative. The domain shift is overcome by pulling positive features closer and pushing negative features further apart. To address class imbalance, the loss function is weighted according to both domain size and class size. In~\cite{gu2022tackling}, the semantic embeddings extracted from class names are utilized to align the visual representations to overcome domain shifts. Furthermore, a domain-specific distribution calibrated loss is introduced to address the class imbalance. However, these methods overlook the fine-grained nature of the datasets. Due to the presence of visually similar categories, aligning visual features of the same classes across domains increases the risk of pulling these food categories (e.g., chicken rice and boiled kampung chicken in Fig.~\ref{fig_fg_samples}) even closer in the feature space, thereby deteriorating fine-grained recognition.

In this paper, we exploit the rich knowledge of LLMs to address challenges in food recognition. First, we leverage LLMs to generate food titles and ingredients. The food titles generally include major ingredients and cooking methods. For example, despite the milk images in Fig.\ref{fig_da_samples} being sampled from different domains, LLMs can generate the same text, ``milk'', for both images. With LLMs-generated texts as the ``bridge'' between domains, cross-modal alignment can effectively bring the cross-domain visual features closer by aligning them with the textual counterparts to overcome the domain gap. The generated ingredients also include minor components, helping to differentiate fine-grained categories with discriminative details. For instance, as shown in Fig.~\ref{fig_fg_samples}, a discriminative region in the ``chicken rice'' image is rice. By supplementing visual embeddings with the LLMs-generated ingredients, such as rice, cucumber and chicken, ``chicken rice'' can be distinguished from other chicken-related dishes, thereby improving fine-grained recognition. Finally, we augment the visual features with textual embeddings for recognition. The textual modality alleviates the under-representation of embeddings caused by insufficient image samples, especially in the tail classes.

We conduct experiments on two food datasets to compare with the existing approaches in domain adaptation~\cite{rangwani2022closer,zhang2023free,zhou2024unsupervised}, imbalance learning~\cite{park2022majority,wang2024unified,du2024probabilistic}, fine-grained recognition~\cite{liu2023learn,min2023large,liang2022penalizing}, long-tailed domain adaptation~\cite{gu2022tackling,yang2022multi}. Our findings include: 
\begin{itemize}
    \item LLMs are capable of generating generic food titles that capture major ingredients in the images. By aligning the visual and textual features in a shared space, a domain-invariant feature space can be effectively learned to alleviate the domain gap. Generating food ingredients is advantageous because they contain unique components specific to each category, which helps produce more separable representations for distinguishing fine-grained categories. Textual modality can supplement visual content more effectively compared to image-mixing augmentation~\cite{park2022majority}.
    \item Our augmentation scheme, when applied to standard networks like ResNet, outperforms well-established methods such as new loss designs~\cite{wang2024unified}, mixtures of experts~\cite{liu2023learn}, and other complex modules~\cite{du2024probabilistic,zhou2024unsupervised} in terms of accuracy on two imbalanced domain-shift food datasets.
\end{itemize}
\section{Related work}
\textbf{Domain Adaptation.} The early unsupervised methods~\cite{ganin2016domain} overcome the domain gaps by training a domain discriminator to learn the domain-invariant features. Variants of methods have been proposed based on this adversarial training, including label smoothing~\cite{zhang2023free} and multi-task learning~\cite{rangwani2022closer}. Recent works~\cite{zhou2024unsupervised} have made attempts to utilize Vision-Language Models to bridge the domain gaps. However, most existing approaches assume a similar balanced dataset distribution across domains, which is rarely the case in the real world. Our work addresses a more practical scenario where datasets are imbalanced across domains and possess distinct categorical distributions. 

\textbf{Long-Tailed Recognition.} Prior research on long-tailed learning focuses on adjusting class-wise contributions by re-sampling~\cite{chawla2002smote}, reweighting~\cite{wang2024unified}, and calibrating logits \cite{ren2020balanced}. The learning paradigms such as contrastive learning~\cite{du2024probabilistic}, and transfer learning~\cite{liu2019large} have also been explored. Data augmentation has also been exploited to over-sample the minority classes~\cite{park2022majority}. However, these existing works assume the datasets sampled from the same domain and ignore the biases induced by domain shifts.

\begin{figure}
    \centering
    \includegraphics[width=0.9\linewidth]{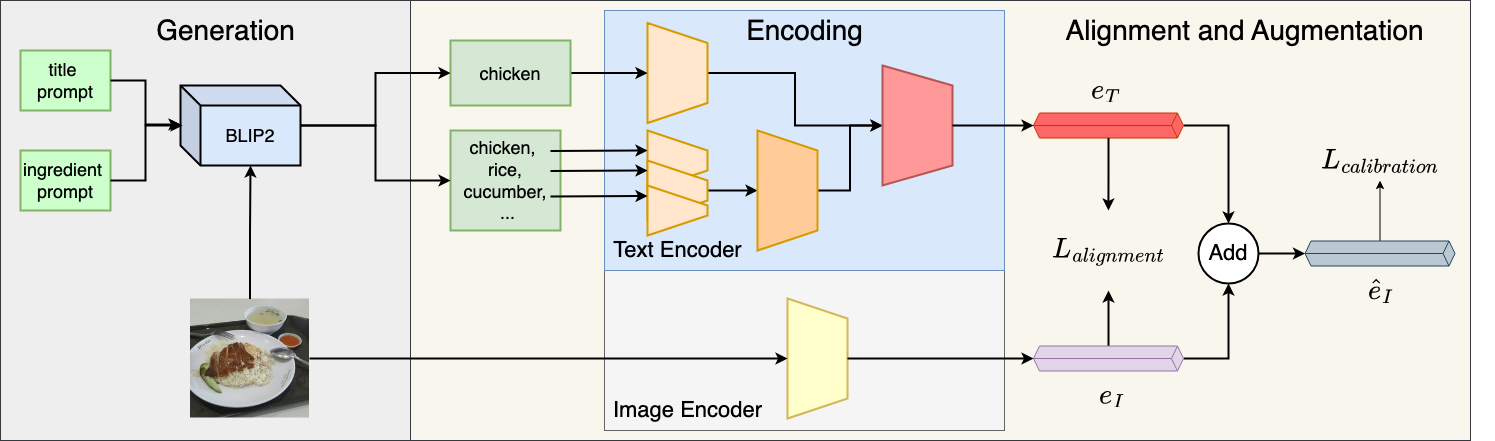}
    \caption{An overview of our proposed framework. The food title and ingredients are generated using LLMs and are encoded by the hierarchical transformers as the embedding $e_{T}$. An image encoder takes the image as input and outputs the visual embedding $e_I$. We align the textual and visual embeddings by enforcing an alignment loss on them and the final embedding is obtained by adding the two modalities. We employ a variant of cross-entropy loss by calibrating the exponentiated logits with their class frequencies to compute the recognition penalties.}
    \label{fig_framework}
\end{figure}

\textbf{Fine-Grained Visual Recognition (FGVR).} The early works~\cite{bossard14} on FGVR aim to identify the key parts of an object and extract part-level features for fine-grained recognition. Another branch of strategies centers on feature learning including crafting loss functions~\cite{liang2022penalizing}, and fusing intermediate features~\cite{liu2023learn,min2023large}. Recently, subset learning~\cite{rodenas2022learning} has been proposed for fine-grained recognition, where the datasets are grouped in multiple subsets. The expert models are trained for each subset to distinguish the classes within that group. In this work, we explore the knowledge of LLMs to generate discriminative texts to distinguish fine-grained classes.

\textbf{Multifaceted Challenges.} The problem of fine-grained domain adaptation is addressed by a framework equipped with an adversarial module for domain alignment and a self-attention module for fine-grained recognition~\cite{wang2020adversarial}. However, in low data regimes, identifying the correct regions for recognition becomes challenging and risks overfitting to spurious correlated patterns from the training data. In~\cite{gu2022tackling}, to tackle class imbalance, a distribution-calibrated classification loss is employed. To address shifts caused by the domain gaps, visual features across domains are aligned closer to their class semantic features extracted from class names. Nonetheless, as some food categories differ by subtle variance in visual appearance, cross-domain visual alignment worsens their classification performance. To improve fine-grained recognition under a long-tailed distribution, attention mechanisms are designed to capture discriminative regions~\cite{shu2022improving}. However, domain gaps can cause a shift in the learned features between training and testing samples, resulting in poor generalization in both imbalanced learning and fine-grained recognition. Unlike previous works, we address all three data challenges simultaneously with LLMs-based text augmentation.


\section{Methodology}

\subsection{Problem Definition}


We are given the training data $\mathcal{D}_s = \{X_{s}^i, Y_{s}^i \}_{i=1}^{n_s}$ with $n_s$ samples drawn from the source distribution $p\left(X_s\right)$. We are also given the training data $\mathcal{D}_t = \{X_{t}^j, Y_{t}^j \}_{j=1}^{n_t}$ of $n_t$ samples drawn from the target distribution $p\left(X_t\right)$. Both domains share the same categories. We assume that there is a covariate shift \cite{shimodaira2000improving} between $X_s$ and $X_t$, that is, there is a difference between the probability distributions $p\left(X_s\right)$ and $p\left(X_t\right)$. Both source and target training datasets are imbalanced and follow different label distributions. Our goal is to learn a model $f$: $X \to Y$ using samples from both domains so that it can perform well on the target domain.

\subsection{Cross-modal Learning}

The overall framework is depicted in Fig.~\ref{fig_framework}. Our aim is to augment visual features with textual labels suggested by LLMs. Specifically, we feed the images along with two prompts to LLMs to generate food titles and ingredients. We then encode the generated texts and images using the hierarchical Transformer~\cite{vaswani2017attention} and ResNet50~\cite{he2016deep}, respectively. We map the embeddings of the two modalities from the source and target domains into a shared feature space and align them using an alignment loss. 


\textbf{Generation} 
We choose BLIP2~\cite{li2023blip} to generate the food titles and ingredients for the images. To generate the title, we feed the image and the following \textbf{title prompt} to the BLIP2: ``Question: What is the food in the image? Answer:''. Similarly, we use the \textbf{ingredient prompt} to generate the ingredients: ``Question: What are the food ingredients in the image? Answer:''. 

\textbf{Encoding} We utilize ResNet-50~\cite{he2016deep} pre-trained on ImageNet as the image encoder. We implement the text encoder as the hierarchical Transformer as done in~\cite{salvador2021revamping}. As shown in Fig.~\ref{fig_framework}, two separate Transformer encoders are used to process the tokenized sentences from the title and ingredients. Since ingredients consist of a sequence of ingredient names, therefore, we utilize a second Transformer encoder to take the list of embeddings as input and output a single embedding for the list of sentences. The text embedding $e_T$ is obtained by projecting the concatenated embeddings from the title and ingredients into the joint image-text embedding space. The text encoder is pre-trained on Recipe1M dataset~\cite{salvador2017learning}. 


\textbf{Alignment and Augmentation} We utilize bi-directional triplet loss~\cite{salvador2021revamping} as the cross-modal alignment loss to pull closer the visual features with their corresponding text features. Given the image-text paired data, we have the alignment loss defined as follows:
\begin{equation}
\begin{split}
\mathcal{L}_{alignment} = \  &  max(0, c({e}_{I}^{a}, e_{T}^{n})-c({e}_{I}^{a}, e_{T}^{p})+\alpha)
\\
 & + max(0, c(e_{T}^{a}, {e}_{I}^{n})-c(e_{T}^{a}, {e}_{I}^{p})+\alpha),
\end{split}
\label{eq:triplet_loss}
\end{equation}
where $c(\cdot)$ is the cosine similarity metric. The superscripts $a$, $p$, and $n$ denote anchor, positive and negative samples, respectively, with the margin $\alpha$ set to 0.3. 


We augment the visual embedding $e_{I}$ with aligned textual embedding $e_{T}$ and obtain the augmented embedding $\hat{e}_{I}$. To address the imbalance in training data, a variant of the cross-entropy loss is utilized~\cite{ren2020balanced}, where the sample logit is normalized by the softmax function first, and then calibrated by its class frequency $n_y$ with the class label being $y$. The calibration-based recognition loss is formulated as follows:

\begin{equation}
    \mathcal{L}_{calibration} = -log\frac{n_ye^{l_y}}{\sum_{i=1}^k{n_i}e^{l_i}},
\end{equation}
where $l_y$ represents the class logit, which is obtained by passing $\hat{e}_I$ through the classifier layer $g$, i.e., $l_y=g(\hat{e}_I)$. The total training object is defined as:

\begin{equation}
    \mathcal{L} = \mathcal{L}_{aligbment} + \mathcal{L}_{calibration}.
\end{equation}

\begin{figure}[t!]\centering
\begin{tabular}{cccccc}
{\scriptsize (a)}\vspace{-0.01cm}& 
{\scriptsize (b)}\vspace{-0.05cm}& 
{\scriptsize (c)}  \vspace{0.05cm}  \\
\makebox[0pt][r]{\makebox[20pt]{\raisebox{45pt}{\rotatebox[origin=c]{90}{ERM}}}}%
\includegraphics[scale=0.15]{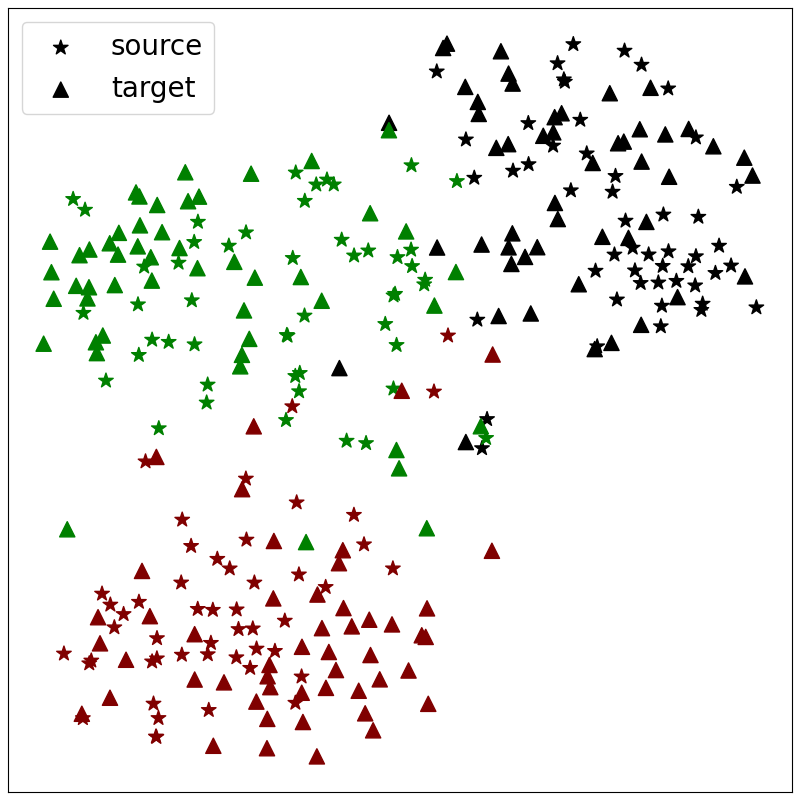} &
\includegraphics[scale=0.15]{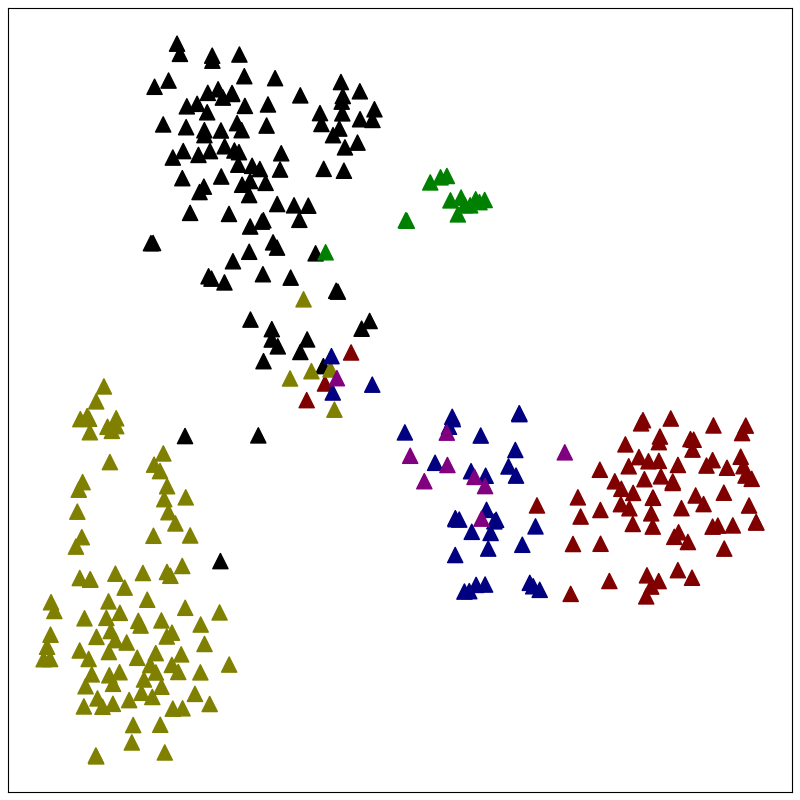} &
\includegraphics[scale=0.15]{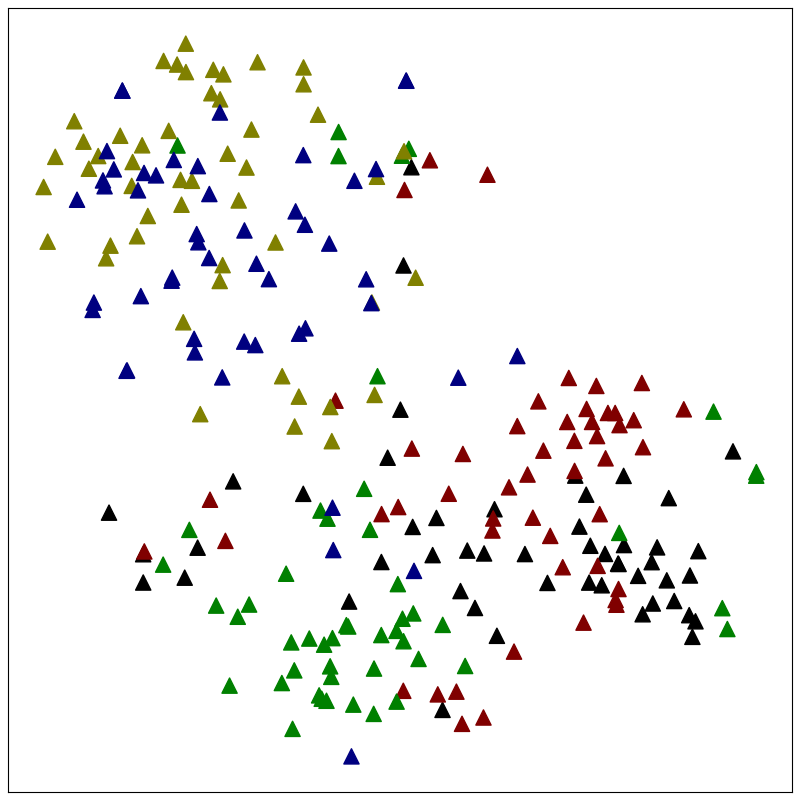} 
\\

\makebox[0pt][r]{\makebox[20pt]{\raisebox{45pt}{\rotatebox[origin=c]{90}{Ours}}}}%
\includegraphics[scale=0.15]{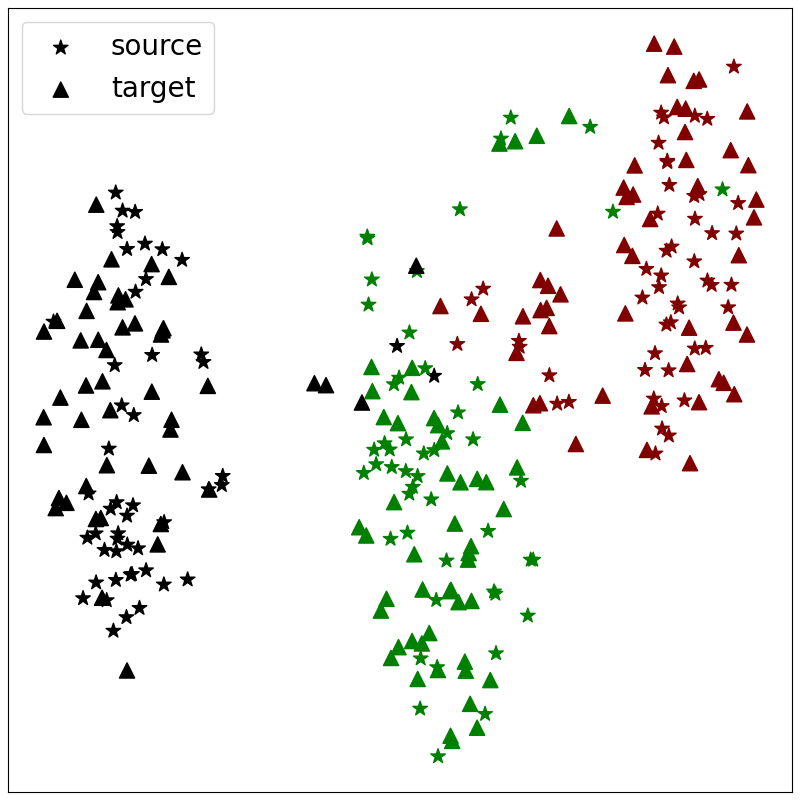} &
\includegraphics[scale=0.15]{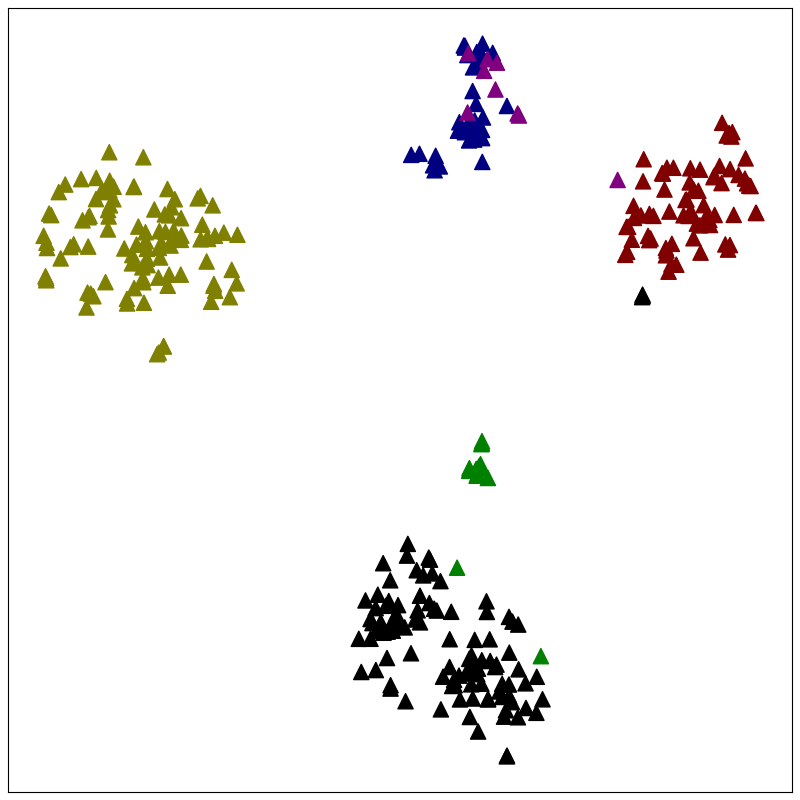} &
\includegraphics[scale=0.15]{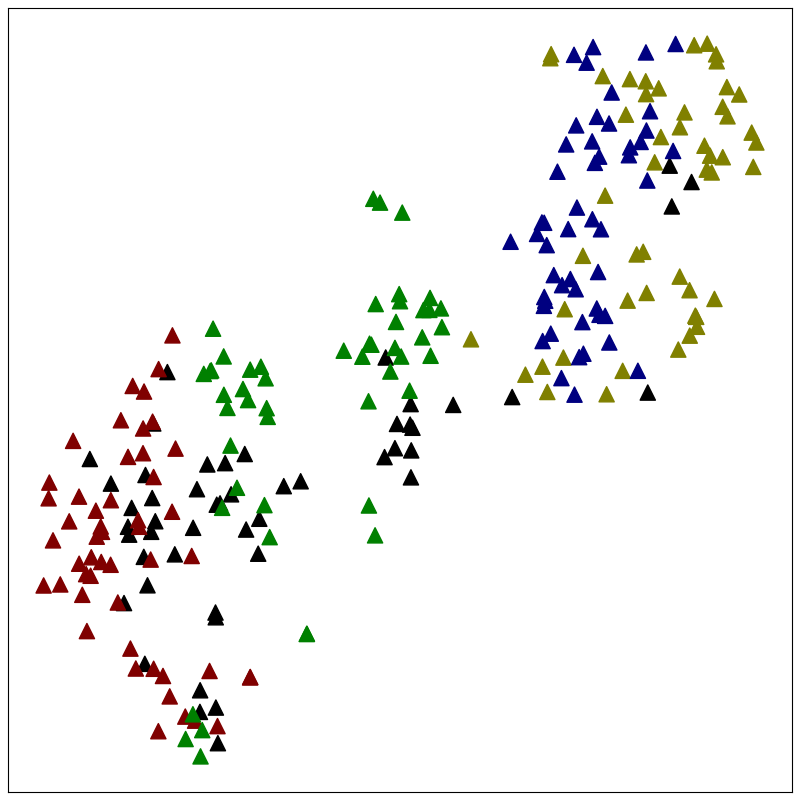} 

\end{tabular}\vspace{-0.1in}
\caption{Visualization of (a) domain alignment, (b) imbalance learning, and (c) fine-grained categories using t-SNE~\cite{van2008visualizing} on FoodAI-HPB. Different colors indicate different classes. The star denotes features from the source domain while the triangle represents features from the target domain.}
\label{fig_visualComapreImage}
\vspace{-0.2in}
\end{figure}

\subsection{Discussions}

To demonstrate the efficacy of our proposed approach in tackling the challenges in the food dataset, we perform t-SNE~\cite{van2008visualizing} visualization on various embeddings learned by the baseline (i.e, ERM~\cite{vapnik1999overview}, which trains the model using cross-entropy loss) and our proposed approach in Fig.~\ref{fig_visualComapreImage}.

As shown in Fig.~\ref{fig_visualComapreImage}a, the ERM-extracted features of source and target domains are spread out and not clustered in the feature space, often mixing with other categories across class boundaries. {In contrast, our approach leverages Eq.~\ref{eq:triplet_loss} to pull images closer to their corresponding texts while pushing them away from unrelated ones. Since LLMs can generate the same titles and similar ingredient descriptions for images within the same categories across domains, aligning images with these texts encourages image features from the same categories to cluster together. This effectively groups cross-domain samples into class-specific regions, leading to tighter feature clusters and a reduced domain gap.}

Incorporating class-relevant semantic information from LLM-generated texts can help learn more separable features for tail classes and address the under-representation caused by insufficient training samples. In Fig.~\ref{fig_visualComapreImage}b, we plot target-domain features from both the head and tail classes. The features extracted by the baseline without any domain alignment present ill-defined decision boundaries, with tail features tending to merge with head ones. In contrast, our approach can separate the head and tail classes with a larger margin.


Compared to identifying discriminative image regions~\cite{bossard14}, using generated text is more effective especially in the case where ingredients occlude each other in the images. Fig.~\ref{fig_visualComapreImage}c visualizes the fine-grained food categories. Due to the inter-class similarities, the baseline is struggling to distinguish fine-grained categories and thus learns a tangled feature space. In contrast, our approach generates more compact feature clusters, resulting in distinct inter-class boundaries.
\section{Experiments}

\begin{table}[t!]
    \caption{Detailed statistics of datasets used in the experiments.}
    \label{tab_stat}\vspace{5pt} 
        \centering
                \resizebox{0.7\linewidth}{!}{

                 \begin{tabular}{ccccccc}
\hline
Domain        & \#Train & \#Test & \#Class              &       \#Head            &        \#Medium           &    \#Tail                \\ \hline
FoodAI        & 364724  & 35850  & \multirow{2}{*}{717} & \multirow{2}{*}{208} & \multirow{2}{*}{234} & \multirow{2}{*}{275} \\ \cline{1-3}
HPB           & 86677   & 3328   &                      &                   &                   &                   \\ \hline
UPMC Food-101 & 15571   & 22716  & \multirow{2}{*}{101} & \multirow{2}{*}{33} & \multirow{2}{*}{34} & \multirow{2}{*}{34} \\ \cline{1-3}
ETH Food-101  & 16454   & 25250  &                      &                   &                   &                   \\ \hline
\end{tabular}
                }
        \label{tab_statistics}
    \end{table}

\subsection{Dataset}

\textbf{FoodAI-HPB.} The FoodAI-HPB dataset comprises two food domains: the source domain (FoodAI~\cite{sahoo2019foodai}), which encompasses instances gathered from the Internet, and the target domain (HPB), which contains food images uploaded by residents in the free-living environment through a mobile app launched by a government agency. As shown in Fig.~\ref{fig:dist}, although both domains showcase long-tailed distributions, they have different imbalanced label distributions. A minority class in one domain might possess sufficient samples in the other domain. The HPB testing dataset is used for evaluation. 

\begin{table}[ht]
    \caption{Performance comparison on \textbf{FoodAI-HPB} dataset with state-of-the-art methods.}
    \label{tab_comparison_food}
    \vspace{4pt}
    \centering
    \resizebox{0.8\linewidth}{!}{
        \begin{tabular}{p{4cm}p{2.2cm}cccccc}
            \hline
            \multirow{2}{*}{} & \multirow{2}{*}{Method} & \multicolumn{4}{c}{Top-1} & Top-5 \\ \cline{3-7} 
                              &                         & Head & Medium & Tail & All & All   \\ \hline
                Baseline              & ERM~\cite{vapnik1999overview}                   &  0.626    &    0.479   &      0.339   &  0.468  &       0.804  \\ \hline
             LLM              & BLIP2~\cite{li2023blip}                   &  0.134    &    0.145   &      0.128   &  0.139  &       0.312  \\ \hline

            \multirow{4}{*}{\raggedright Domain adaptation (DA)} & DANN~\cite{ganin2016domain}                     &   0.606   &   0.515    &     0.424  &  0.507    &     0.684       \\ \cline{2-7} 
                              & ELS~\cite{zhang2023free}       &  0.676   &   0.535   &   0.450    &   0.542        &   0.683      \\ \cline{2-7} 
                              & SDAT~\cite{rangwani2022closer}       &   \textbf{0.686} &  0.531    &   0.448    &    0.544       &  0.685   \\ \cline{2-7} 
                              & VLPUDA~\cite{zhou2024unsupervised}                 &  0.677   &   0.511   &   0.355    &     0.500      &  0.652        \\ \hline
            \multirow{4}{*}{\raggedright Long tailed (LT)}        & BLMS~\cite{ren2020balanced}        &  0.617   &    0.513  &     0.409  &     0.503      &    0.813   \\ \cline{2-7} 
                              & CMO~\cite{park2022majority}                   &   0.529   &    0.474    &   0.399  &  0.461   &   0.761    \\ \cline{2-7} 
                              & VS~\cite{wang2024unified}           &   0.596   &   0.510    &    0.430    &   0.504   &   0.814          \\ \cline{2-7} 
                              & ProCo~\cite{du2024probabilistic}         &  0.610   &   0.513   &    0.420   &    0.505       &   0.821      \\ \hline
            \multirow{4}{*}{\raggedright Fine-grained (FG)}  & PARNet~\cite{qiu2022mining}    &    0.641           &   0.489   &    0.346    &   0.478  &  0.804        \\ \cline{2-7} 
                              & CMAL~\cite{liu2023learn}         &  0.673     &   0.532    &    0.468       &       0.548     & 0.824      \\ \cline{2-7} 
                              & PRENet~\cite{min2023large}        &   0.655  &   0.507   &    0.367   &      0.503     &   0.815      \\ \cline{2-7} 
                              & MHEM~\cite{liang2022penalizing}                        &  0.656    &     0.507   &  0.427   &   0.519  &  0.848     \\ \hline
            \multirow{2}{*}{\raggedright DA+LT}         & LTDS~\cite{gu2022tackling}       &   0.651  &  0.580    &    0.545   &    0.587       &    0.863       \\ \cline{2-7} 
                              & BoDA~\cite{yang2022multi}         &   0.646  &   0.487   &  0.342     &    0.477       &     0.789           \\ \hline
               DA+LT+FG               & Ours            &   0.661  &    \textbf{0.602}  &   \textbf{0.572}    &    \textbf{0.607}       &   \textbf{0.872}       \\ \hline
        \end{tabular}
    }
\end{table}

\textbf{UPMC-ETH Food101} We construct a two-domain food dataset for experimental evaluation. We adopt UPMC Food-101 \cite{wang2015recipe} as the source domain where samples are collected from the web. To introduce long-tailed characteristics, we implement an exponential decay strategy in the number of training instances for each food category. Likewise, we construct an imbalanced target domain by utilizing the ETH Food-101 dataset~\cite{bossard14}, where food images are uploaded by users.  In this process, we shuffle the food categories to generate a distinct label distribution from that of UPMC. To gauge the extent of imbalance, we employ the imbalance ratio as a quantifying metric. The imbalance ratio is defined as the ratio between the sample sizes of the most frequent and least frequent classes within the dataset. We keep the original balanced ETH Food-101 test dataset and use it as our target test dataset. It's noteworthy that both the UPMC and ETH Food-101 datasets exhibit an identical imbalance ratio of 100. The detailed dataset statistics can be found in Table.~\ref{tab_statistics}.

\textbf{Implementation details}
 We use ResNet-50 \cite{he2016deep} pre-trained on ImageNet as the backbone network. The text encoder is implemented as Transformers~\cite{vaswani2017attention} and pre-trained on Recipe1M dataset~\cite{salvador2017learning}.  {During training,  we freeze the LLMs and jointly train both encoders using the SGD optimizer with a learning rate of 0.001.} The batch size is set as 128 for each domain.
 
\textbf{Competing Methods}
We compare our method with different algorithms that span various learning approaches and categories, including (1) baseline: ERM~\cite{vapnik1999overview}, which represents the model trained using cross-entropy loss, (2) LLM: BLIP2~\cite{li2023blip}, where the similarity between the images and each of the class names is computed, and then the class with the highest similarity score is selected as the prediction, (3) domain adaptation learning: DANN~\cite{ganin2016domain}, SDAT~\cite{rangwani2022closer}, ELS~\cite{zhang2023free}, VLPUDA~\cite{zhou2024unsupervised}, (4) imbalance learning: BLMS~\cite{ren2020balanced}, CMO~\cite{park2022majority}, VS~\cite{wang2024unified},  ProCo~\cite{du2024probabilistic}, (5) fine-grained recognition: PARNet~\cite{qiu2022mining}, CMAL~\cite{liu2023learn}, PRENet~\cite{min2023large},  MHEM~\cite{liang2022penalizing}, (6) long-tailed domain adaptation leaning: LTDS~\cite{gu2022tackling}, BoDA~\cite{yang2022multi}.

\begin{table}[t]
    \caption{Performance comparison on \textbf{UPMC-ETH Food101}.}
\label{tab_comparison_food101}\vspace{5pt}
    \centering
   \resizebox{0.8\linewidth}{!}{
    \begin{tabular}{p{4cm}p{2.2cm}cccccc}
\hline
\multirow{2}{*}{}             & \multirow{2}{*}{Method} & \multicolumn{4}{c}{Top-1} & Top-5 \\ \cline{3-7} 
                              &                         & Head & Medium & Tail & All & All   \\ \hline
              Baseline                & ERM~\cite{vapnik1999overview}                &   0.845  &   0.555   &    0.277   &       0.556    &      0.819          \\ \hline
                              LLM              & BLIP2~\cite{li2023blip}                   &  0.672    &    0.704  &      0.639   &  0.672  &       0.921  \\ \hline
\multirow{4}{*}{\raggedright Domain adaptation (DA)}           &  DANN~\cite{ganin2016domain}                   &   0.836  &   0.546   &   0.315    &     0.563      &       0.689        \\ \cline{2-7} 
                              &  ELS~\cite{zhang2023free}       &   0.834  &   0.667   &  0.525     &    0.674       &   0.783  \\ \cline{2-7} 
                              &   SDAT~\cite{rangwani2022closer}          &  0.823   &   0.674   &   0.508    &     0.667      &     0.781      \\ \cline{2-7} 
                              &   VLPUDA~\cite{zhou2024unsupervised}                 &  0.880   &  0.777    &    0.604   &    0.752       &     0.868      \\ \hline
\multirow{4}{*}{\raggedright Long tailed (LT)}           &  BLMS~\cite{ren2020balanced}       &  0.803   &   0.682   &    0.454   &   0.645        &   0.884        \\ \cline{2-7} 
                              &   CMO~\cite{park2022majority}                      &  0.847    &    0.721    &  0.548   &   0.704  &  0.892     \\ \cline{2-7} 
                              & VS~\cite{wang2024unified}     &   0.786  &  0.681    &   0.437    &     0.633      &       0.875        \\ \cline{2-7} 
                              &  ProCo~\cite{du2024probabilistic}          &  0.814    &    0.712  &   0.493    &     0.672      &    0.895       \\ \hline
\multirow{4}{*}{\raggedright Fine-grained (FG)} &  PARNet~\cite{qiu2022mining}      &  0.842   &   0.601     &  0.330       &    0.589 & 0.855      \\ \cline{2-7} 
                              &  CMAL~\cite{liu2023learn}         &  0.875     &   0.721    &    0.515       &       0.702     & 0.914      \\ \cline{2-7} 
                              &    PRENet~\cite{min2023large}                  & 0.863    &   0.661   &    0.426   &     0.648      &  0.874           \\ \cline{2-7} 
                              &   MHEM~\cite{liang2022penalizing}          &  0.819      &  0.603   &   0.439  &   0.618 & 0.848    \\ \hline
\multirow{2}{*}{\raggedright DA+LT}         &  BoDA~\cite{yang2022multi}     &   0.820  &  0.538    &    0.329   &     0.560      &  0.825      \\ \cline{2-7} 
                              &     LTDS~\cite{gu2022tackling}     &  0.800   &   0.689   &  0.491     &    0.659       &    0.900                \\ \hline
DA+LT+FG & Ours           &  \textbf{ 0.886 } &  \textbf{ 0.874 } &  \textbf{ 0.808  }  &    \textbf{0.856 }      &  \textbf{ 0.969  }        \\ \hline
\end{tabular}
   }
\end{table}
\textbf{Evaluation Metric}
Following the standard protocol of \cite{liu2019large}, besides the Top-1 and Top-5 classification accuracy, we also report the accuracy of three disjoint groups based on the label distribution of the target domain: head classes (classes each with more than 70 training instances), medium classes (classes each with 15$\sim$70 training instances), and tail classes (classes under 15 training instances). Table.~\ref{tab_statistics} shows the number of classes per group.

\subsection{Comparison to Existing Works}
The results of all the evaluated methods on the FoodAI-HPB and UPMC-ETH Food101 datasets are presented in Table~\ref{tab_comparison_food} and Table~\ref{tab_comparison_food101}, respectively. The best-performing results in each column are highlighted. DA generally offers higher Top-1 but lower Top-5 accuracies compared to LT methods on both datasets. This is due to the mixing of feature spaces for visually similar categories after domain alignment, which results in fine-grained categories appearing within the Top-5 predictions and causing lower Top-5 accuracies. Imbalance learning methods typically involve a performance trade-off, enhancing accuracies across medium-shot and few-shot classes at the expense of many-shot classes. Specifically, CMO uses an image-mixing augmentation technique, i.e., CutMix. However, this augmentation fails on more challenging datasets like FoodAI-HPB due to the presence of many fine-grained categories, where image-mixing blurs the boundaries of these categories. By combining DA and LT, LTDS effectively alleviates the problems caused by DA or LT alone, thereby improving performance across all metrics on FoodAI-HPB. This is due to the effective domain alignment, which brings visual features closer to the corresponding class semantic features while reweighting the recognition loss to address data imbalance.

\begin{figure}
    \centering
    \includegraphics[width=0.57\linewidth]{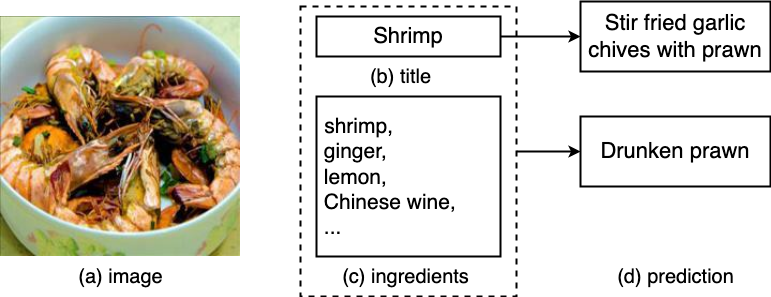}
    \caption{One example providing insights on text augmentation. The correct food class of the input image is ``drunken prawn''. By using only visual features (a), the predicted category is ``steamed crab''. By augmenting with LLM-generated title (b), a better prediction ``stir fried garlic chives with prawn'' is obtained. If both the title and ingredients are augmented (c), the correct food class is predicted.}
    \label{fig_title_ing_aug_help_one}
\end{figure}

FG-based methods similarly improve all metrics due to the learned discriminative features. For example, both CMAL and PRENet utilize a mixture-of-experts approach, maintaining multiple network branches to extract complementary multi-scale feature representations and fuse these features into a unified representation. This helps identify discriminative region features for fine-grained recognition and utilizes diversified features from different expert branches for imbalanced learning. 
We also compare our approach to LLM-based recognition using BLIP2. On the UPMC-ETH Food101 dataset, BLIP2 performs competitively, achieving the second-best Top-5 accuracy of 92.1\%. However, BLIP2 fails on the challenging FoodAI-HPB dataset, showing a significant performance gap compared to other methods, with Top-1 and Top-5 accuracies of 13.9\% and 31.2\%, respectively. We attribute this to the presence of a large number of geography-specific foods, which BLIP2 fails to capture during the model pre-training, leading to unsatisfactory image-text matching.

In contrast, our proposed approach achieves the best Top-1 and Top-5 accuracy over all other methods on both datasets. Moreover, our approach significantly elevates the performance of all three class groups, especially in medium and tail classes. Notably, on UPMC-ETH Food101, an improvement of 53\% for tail classes is achieved in comparison to the ERM baseline. It is worth noting that the proposed approach introduces minimal computational overhead. Typically, BLIP2 generates a short title and an average of four ingredients for each food image. Encoding these textual representations incurs almost no overhead.


\begin{table}[t]
        \vspace{5pt}
        \centering
        \caption{Ablation studies on the components used in the proposed method on FoodAI-HPB. The performance of the first row (without any ticks) refers to the baseline trained with cross-entropy loss. If $L_{alignment}$ is checked, we apply alignment loss during the model training. By incorporating the title texts, we augment the visual features with the textual features of titles.}
        \resizebox{0.8\linewidth}{!}{
            \begin{tabular}{ccccccccc}
\hline
\multirow{2}{*}{$L_{alignment}$ } & \multirow{2}{*}{$L_{calibration}$ } & \multirow{2}{*}{Title} & \multirow{2}{*}{Ingredients} & \multicolumn{4}{c}{Top-1} & Top-5 \\ \cline{5-9} 
               &                    &                     &      & Head    & Medium    & Tail  & All  & All    \\ \hline
                                 &       &                       &       &  0.626    &    0.479   &      0.339   &  0.468  &       0.804 \\ \hline

           \checkmark           &       &                       &      &  0.681            &   0.556   & 0.409 &  0.536  &  0.851      \\ \hline
                \checkmark    &            \checkmark     & &       &   0.649      &   0.591    & 0.547      & 0.587 & 0.860   \\ \hline
           \checkmark           &            \checkmark           &             \checkmark           &                     &   0.650   &   0.600   &  0.559    &  0.599   &   0.868   \\ \hline
             \checkmark         &        \checkmark               &              \checkmark          &            \checkmark           &   0.661  &    0.602  &   0.572    &    0.607       &   0.872   \\ \hline
\end{tabular}
        }
        \label{tab_compo}
    \end{table}




\begin{table}
\caption{Comparison of different LLMs.} 

\begin{subtable}[c]{0.50\textwidth}
\centering
\subcaption{FoodAI-HPB}
\begin{adjustbox}{width=0.95\textwidth}
     \begin{tabular}{cccccc}
\hline
\multirow{2}{*}{} & \multicolumn{4}{c}{Top-1}  & Top-5 \\ \cline{2-6} 
                 & Head & Medium & Tail & All & All \\ \hline
BLIP2             & 0.661    &  0.602    &  0.572      &   0.607  &   0.868   \\ \hline
LLaVA           &  0.661   &  0.594    &   0.559     &  0.600   &  0.866    \\ \hline
FoodLMM           &  0.647   &  0.600    &   0.557     &  0.597   &  0.862    \\ \hline

\end{tabular}
    \end{adjustbox}
\end{subtable} 
\begin{subtable}[c]{0.50\textwidth}
\centering
\subcaption{UPMC-ETH Food101}
 \begin{adjustbox}{width=0.95\textwidth}
     \begin{tabular}{cccccc}
\hline
\multirow{2}{*}{} & \multicolumn{4}{c}{Top-1}  & Top-5 \\ \cline{2-6} 
                   & Head & Medium & Tail & All & All \\ \hline
BLIP2             &  0.886   &   0.874   &  0.808      &   0.856  &   0.969   \\ \hline
LLaVA           &  0.847   &  0.810    &   0.730     &  0.795   &  0.953    \\ \hline
FoodLMM           &  0.842   &   0.798   &    0.730    &   0.789  &  0.946    \\ \hline

\end{tabular}
    \end{adjustbox}
\end{subtable}
\label{tab_llm_choice}
\end{table}

An example is shown in Fig.~\ref{fig_title_ing_aug_help_one} to give insights into how textual augmentation helps food recognition. The generated title is ``shrimp''. Augmenting the visual feature with the title alone is helpful but insufficient to distinguish various shrimp-related dishes, resulting in an incorrect prediction of ``stir fried garlic chives with prawn''. As BLIP2 can generate minor ingredients like ``Chinese wine'' and ``ginger'', these ingredients provide fine-grained details enough for correct classification of the image as ``drunken prawn''.




\subsection{Ablation Analysis}

\subsubsection{Contribution of Components}
Table~\ref{tab_compo} presents the effectiveness of each component used in our approach. By aligning the visual features and their texts with the alignment loss, we encourage the samples from the same categories but in different domains to be closer and alleviate the domain gaps, resulting in a large Top-1 accuracy improvement from 0.468 to 0.536. Utilizing the calibration loss to overcome the data imbalance will push the Top-1 accuracy to 0.587. Further augmenting the visual features with the title keeps boosting the recognition performance, with an extra gain of 1.2\% in Top-1 accuracy. With both title and ingredients augmentation, the best Top-1 accuracy is achieved, i.e., 0.607.

\subsubsection{Choices of LLMs}
Besides BLIP2, we also explore LLaVA~\cite{liu2024visual} and FoodLMM~\cite{yin2023foodlmm} for text generation and compare their impacts on the model performance. The results are presented in Table~\ref{tab_llm_choice}. On both datasets, regardless of what LLMs are being utilized, text augmentation improves all recognition metrics compared to the baseline. Both LLaVA and FoodLMM achieve similar performance because FoodLMM is built on LLaVA and trained on multiple public food datasets for multi-task learning using the instruct-following paradigm. Compared to FoodLMM, the texts generated by BLIP2 are more generic, consisting mainly of ingredients and cooking methods. FoodLMM embeds additional details such as taste (e.g., ``sweet and sour spareribs'') and place of origin (e.g., ``Indian chicken curry'') into the food title, which makes cross-modal alignment challenging and deteriorates the results.

\subsection{Group-wise Performance}
        
        

\begin{figure}[t!]\centering
\begin{tabular}{cccccc}

\includegraphics[scale=0.33]{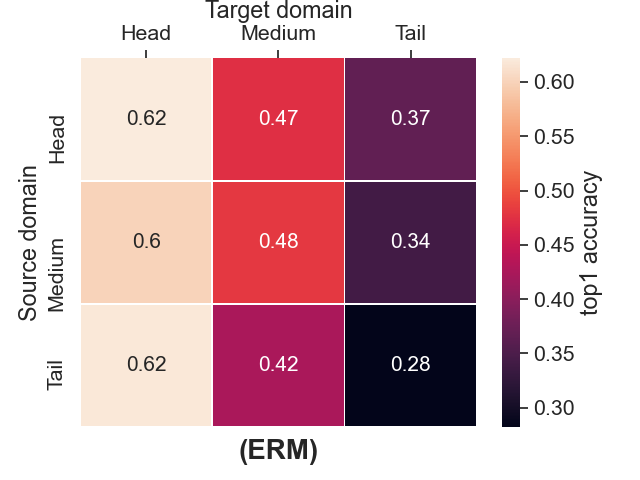} &
\includegraphics[scale=0.33]{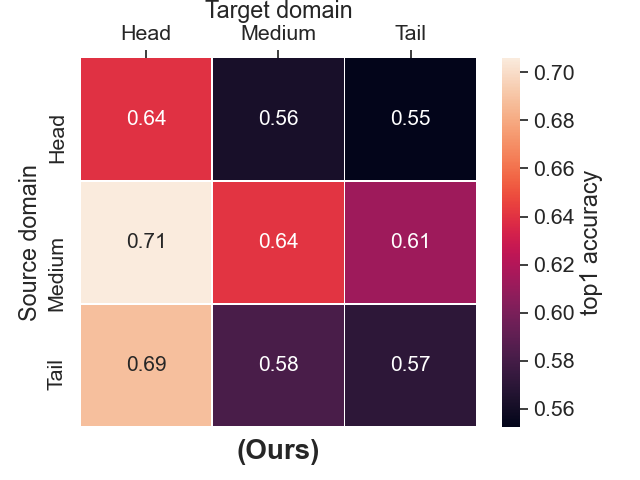} 
\\
\end{tabular}\vspace{-0.1in}
\caption{Top-1 accuracies of (a) the baseline method, ERM, (b) our proposed approach for different class groups of source and target domains on FoodAI-HPB.}
        \label{fig_g2g}
\vspace{-0.2in}
\end{figure}
We further divide the classes into nine disjoint groups based on the label distribution of the source and target domains, such as head-head classes (classes with ample samples in both source and target domains) and head-tail classes (classes with many samples in the source domain but few in the target domain). As depicted in Fig.~\ref{fig_g2g}, our approach introduces substantial improvement for all the nine disjoint groups. The improvement for the tail classes is especially significant, ranging from 18\% to 29\% compared to ERM. It is worth noticing that, however, the number of samples in the source domain appears to have no apparent effect on the performance. While this finding is surprising, we attribute this to the fact that the LLM-generated texts for source domain training examples are mostly consistent. Hence, even with much less training examples, tail classes will not be impacted negatively so long as the generated texts are correct.

\section{Conclusions}



We have presented a simple approach leveraging LLMs to alleviate several challenges common in the real-world food datasets. With LLMs-generated titles and ingredients as supplements to visual features to align images of different domains, cross-modal learning can be more effectively carried out to amend the class boundaries of tail classes while not sacrificing the accuracy of fine-grained classification. While the results are encouraging, the current approach can be impacted negatively by the incorrect generation of texts, especially the food title. Future work includes validation of LLM generation for selective augmentation of texts for food recognition.

\section{Acknowledge}

This research/project is supported by the Ministry of Education, Singapore, under its Academic Research Fund Tier 2 (Proposal ID: T2EP20222-0046). Any opinions, findings and conclusions or recommendations expressed in this material are those of the authors and do not reflect the views of the Ministry of Education, Singapore.

%
%
%
\bibliographystyle{splncs04}
\bibliography{mybibliography}

@article{shimodaira2000improving,
  title={Improving predictive inference under covariate shift by weighting the log-likelihood function},
  author={Shimodaira, Hidetoshi},
  journal={Journal of statistical planning and inference},
  volume={90},
  number={2},
  pages={227--244},
  year={2000},
  publisher={Elsevier}
}

@article{ganin2016domain,
  title={Domain-adversarial training of neural networks},
  author={Ganin, Yaroslav and Ustinova, Evgeniya and Ajakan, Hana and Germain, Pascal and Larochelle, Hugo and Laviolette, Fran{\c{c}}ois and Marchand, Mario and Lempitsky, Victor},
  journal={The journal of machine learning research},
  volume={17},
  number={1},
  pages={2096--2030},
  year={2016},
  publisher={JMLR. org}
}

@inproceedings{yang2022multi,
  title={On multi-domain long-tailed recognition, imbalanced domain generalization and beyond},
  author={Yang, Yuzhe and Wang, Hao and Katabi, Dina},
  booktitle={European Conference on Computer Vision},
  pages={57--75},
  year={2022},
  organization={Springer}
}

@inproceedings{liu2019large,
  title={Large-scale long-tailed recognition in an open world},
  author={Liu, Ziwei and Miao, Zhongqi and Zhan, Xiaohang and Wang, Jiayun and Gong, Boqing and Yu, Stella X},
  booktitle={Proceedings of the IEEE/CVF Conference on Computer Vision and Pattern Recognition},
  pages={2537--2546},
  year={2019}
}

@inproceedings{kang2019decoupling,
  title={Decoupling Representation and Classifier for Long-Tailed Recognition},
  author={Kang, Bingyi and Xie, Saining and Rohrbach, Marcus and Yan, Zhicheng and Gordo, Albert and Feng, Jiashi and Kalantidis, Yannis},
  booktitle={International Conference on Learning Representations},
  year={2019}
}

@article{ren2020balanced,
  title={Balanced meta-softmax for long-tailed visual recognition},
  author={Ren, Jiawei and Yu, Cunjun and Ma, Xiao and Zhao, Haiyu and Yi, Shuai and others},
  journal={Advances in neural information processing systems},
  volume={33},
  pages={4175--4186},
  year={2020}
}

@inproceedings{he2016deep,
  title={Deep residual learning for image recognition},
  author={He, Kaiming and Zhang, Xiangyu and Ren, Shaoqing and Sun, Jian},
  booktitle={Proceedings of the IEEE conference on computer vision and pattern recognition},
  pages={770--778},
  year={2016}
}

@article{vapnik1999overview,
  title={An overview of statistical learning theory},
  author={Vapnik, Vladimir N},
  journal={IEEE transactions on neural networks},
  volume={10},
  number={5},
  pages={988--999},
  year={1999},
  publisher={IEEE}
}

@inproceedings{bossard14,
  title = {Food-101 -- Mining Discriminative Components with Random Forests},
  author = {Bossard, Lukas and Guillaumin, Matthieu and Van Gool, Luc},
  booktitle = {European Conference on Computer Vision},
  year = {2014}
}

@inproceedings{wang2015recipe,
  title={Recipe recognition with large multimodal food dataset},
  author={Wang, Xin and Kumar, Devinder and Thome, Nicolas and Cord, Matthieu and Precioso, Frederic},
  booktitle={2015 IEEE International Conference on Multimedia \& Expo Workshops (ICMEW)},
  pages={1--6},
  year={2015},
  organization={IEEE}
}

@inproceedings{gu2022tackling,
  title={Tackling long-tailed category distribution under domain shifts},
  author={Gu, Xiao and Guo, Yao and Li, Zeju and Qiu, Jianing and Dou, Qi and Liu, Yuxuan and Lo, Benny and Yang, Guang-Zhong},
  booktitle={Computer Vision--ECCV 2022: 17th European Conference, Tel Aviv, Israel, October 23--27, 2022, Proceedings, Part XXIII},
  pages={727--743},
  year={2022},
  organization={Springer}
}

@article{yin2023foodlmm,
  title={Foodlmm: A versatile food assistant using large multi-modal model},
  author={Yin, Yuehao and Qi, Huiyan and Zhu, Bin and Chen, Jingjing and Jiang, Yu-Gang and Ngo, Chong-Wah},
  journal={arXiv preprint arXiv:2312.14991},
  year={2023}
}

@inproceedings{salvador2021revamping,
  title={Revamping cross-modal recipe retrieval with hierarchical transformers and self-supervised learning},
  author={Salvador, Amaia and Gundogdu, Erhan and Bazzani, Loris and Donoser, Michael},
  booktitle={Proceedings of the IEEE/CVF Conference on Computer Vision and Pattern Recognition},
  pages={15475--15484},
  year={2021}
}

@article{van2008visualizing,
  title={Visualizing data using t-SNE.},
  author={Van der Maaten, Laurens and Hinton, Geoffrey},
  journal={Journal of machine learning research},
  volume={9},
  number={11},
  year={2008}
}

@inproceedings{rangwani2022closer,
  title={A closer look at smoothness in domain adversarial training},
  author={Rangwani, Harsh and Aithal, Sumukh K and Mishra, Mayank and Jain, Arihant and Radhakrishnan, Venkatesh Babu},
  booktitle={International conference on machine learning},
  pages={18378--18399},
  year={2022},
  organization={PMLR}
}

@article{zhang2023free,
  title={Free lunch for domain adversarial training: Environment label smoothing},
  author={Zhang, YiFan and Wang, Xue and Liang, Jian and Zhang, Zhang and Wang, Liang and Jin, Rong and Tan, Tieniu},
  journal={arXiv preprint arXiv:2302.00194},
  year={2023}
}

@article{zhou2024unsupervised,
  title={Unsupervised Domain Adaption Harnessing Vision-Language Pre-training},
  author={Zhou, Wenlve and Zhou, Zhiheng},
  journal={IEEE Transactions on Circuits and Systems for Video Technology},
  year={2024},
  publisher={IEEE}
}

@article{wang2024unified,
  title={A unified generalization analysis of re-weighting and logit-adjustment for imbalanced learning},
  author={Wang, Zitai and Xu, Qianqian and Yang, Zhiyong and He, Yuan and Cao, Xiaochun and Huang, Qingming},
  journal={Advances in Neural Information Processing Systems},
  volume={36},
  year={2024}
}

@article{du2024probabilistic,
  title={Probabilistic contrastive learning for long-tailed visual recognition},
  author={Du, Chaoqun and Wang, Yulin and Song, Shiji and Huang, Gao},
  journal={IEEE Transactions on Pattern Analysis and Machine Intelligence},
  year={2024},
  publisher={IEEE}
}

@article{qiu2022mining,
  title={Mining discriminative food regions for accurate food recognition},
  author={Qiu, Jianing and Lo, Frank P-W and Sun, Yingnan and Wang, Siyao and Lo, Benny},
  journal={arXiv preprint arXiv:2207.03692},
  year={2022}
}

@article{liu2023learn,
  title={Learn from each other to classify better: Cross-layer mutual attention learning for fine-grained visual classification},
  author={Liu, Dichao and Zhao, Longjiao and Wang, Yu and Kato, Jien},
  journal={Pattern Recognition},
  volume={140},
  pages={109550},
  year={2023},
  publisher={Elsevier}
}

@article{min2023large,
  title={Large scale visual food recognition},
  author={Min, Weiqing and Wang, Zhiling and Liu, Yuxin and Luo, Mengjiang and Kang, Liping and Wei, Xiaoming and Wei, Xiaolin and Jiang, Shuqiang},
  journal={IEEE Transactions on Pattern Analysis and Machine Intelligence},
  volume={45},
  number={8},
  pages={9932--9949},
  year={2023},
  publisher={IEEE}
}

@article{liang2022penalizing,
  title={Penalizing the hard example but not too much: A strong baseline for fine-grained visual classification},
  author={Liang, Yuanzhi and Zhu, Linchao and Wang, Xiaohan and Yang, Yi},
  journal={IEEE Transactions on Neural Networks and Learning Systems},
  year={2022},
  publisher={IEEE}
}

@inproceedings{salvador2017learning,
  title={Learning cross-modal embeddings for cooking recipes and food images},
  author={Salvador, Amaia and Hynes, Nicholas and Aytar, Yusuf and Marin, Javier and Ofli, Ferda and Weber, Ingmar and Torralba, Antonio},
  booktitle={Proceedings of the IEEE conference on computer vision and pattern recognition},
  pages={3020--3028},
  year={2017}
}

@article{vaswani2017attention,
  title={Attention is all you need},
  author={Vaswani, Ashish and Shazeer, Noam and Parmar, Niki and Uszkoreit, Jakob and Jones, Llion and Gomez, Aidan N and Kaiser, {\L}ukasz and Polosukhin, Illia},
  journal={Advances in neural information processing systems},
  volume={30},
  year={2017}
}

@article{chawla2002smote,
  title={SMOTE: synthetic minority over-sampling technique},
  author={Chawla, Nitesh V and Bowyer, Kevin W and Hall, Lawrence O and Kegelmeyer, W Philip},
  journal={Journal of artificial intelligence research},
  volume={16},
  pages={321--357},
  year={2002}
}

@inproceedings{park2022majority,
  title={The majority can help the minority: Context-rich minority oversampling for long-tailed classification},
  author={Park, Seulki and Hong, Youngkyu and Heo, Byeongho and Yun, Sangdoo and Choi, Jin Young},
  booktitle={Proceedings of the IEEE/CVF conference on computer vision and pattern recognition},
  pages={6887--6896},
  year={2022}
}

@inproceedings{rodenas2022learning,
  title={Learning multi-subset of classes for fine-grained food recognition},
  author={R{\'o}denas, Javier and Nagarajan, Bhalaji and Bola{\~n}os, Marc and Radeva, Petia},
  booktitle={Proceedings of the 7th International Workshop on Multimedia Assisted Dietary Management},
  pages={17--26},
  year={2022}
}

@inproceedings{li2023blip,
  title={Blip-2: Bootstrapping language-image pre-training with frozen image encoders and large language models},
  author={Li, Junnan and Li, Dongxu and Savarese, Silvio and Hoi, Steven},
  booktitle={International conference on machine learning},
  pages={19730--19742},
  year={2023},
  organization={PMLR}
}

@inproceedings{wang2020adversarial,
  title={An adversarial domain adaptation network for cross-domain fine-grained recognition},
  author={Wang, Yimu and Song, Renjie and Wei, Xiu-Shen and Zhang, Lijun},
  booktitle={Proceedings of the IEEE/CVF Winter Conference on Applications of Computer Vision},
  pages={1228--1236},
  year={2020}
}

@inproceedings{shu2022improving,
  title={Improving fine-grained visual recognition in low data regimes via self-boosting attention mechanism},
  author={Shu, Yangyang and Yu, Baosheng and Xu, Haiming and Liu, Lingqiao},
  booktitle={European Conference on Computer Vision},
  pages={449--465},
  year={2022},
  organization={Springer}
}

@article{liu2024visual,
  title={Visual instruction tuning},
  author={Liu, Haotian and Li, Chunyuan and Wu, Qingyang and Lee, Yong Jae},
  journal={Advances in neural information processing systems},
  volume={36},
  year={2024}
}

@inproceedings{sahoo2019foodai,
  title={FoodAI: Food image recognition via deep learning for smart food logging},
  author={Sahoo, Doyen and Hao, Wang and Ke, Shu and Xiongwei, Wu and Le, Hung and Achananuparp, Palakorn and Lim, Ee-Peng and Hoi, Steven CH},
  booktitle={Proceedings of the 25th ACM SIGKDD international conference on knowledge discovery \& data mining},
  pages={2260--2268},
  year={2019}
}

\end{document}